# Two Algorithms for Deciding Coincidence In Double Temporal Recurrence of Eventuality Sequences

Babatunde Opeoluwa Akinkunmi[a] and Adesoji A. Adegbola[b]
Dept of Computer Science,
University of Ibadan, Ibadan,
Nigeria

[a]email: ope34648@yahoo.com
[b]email: soji.adegbola@google.com

**Abstract**
Let two disjoint sequences of eventualities x (signifying the sequence $x_0$, $x_1$, $x_2 \ldots x_{n-1}$) and y (signifying the sequence $y_0$, $y_1$, $y_2 \ldots y_{m-1}$) both recur over the same time interval (i.e. double recurrence (x, y)) and it is required to determine whether or not a subinterval exists within the said interval, which is a common subinterval of the intervals of occurrence of some $x_p$ and $y_q$.

This paper presents two algorithms for solving the problem. The first explores an arbitrary cycle of the double recurrence for the existence of such an interval. Its worst case running time is quadratic.

The other algorithm is based on the novel notion of gcd-partitions and has a linear worst case running time. If eventuality sequence pair (w, z) is a gcd partition for the double recurrence (x, y), then, from a certain property of gcd-partitions, within any cycle of the double recurrence, there exists r and s such that intervals of occurrence of each of $x_p$ and $y_q$ are non-disjoint with the interval of co-occurrence of $w_r$ and $z_s$. As such, a coincidence between $x_p$ and $y_q$ occurs within a cycle of double recurrence if and only if such r and s exist so that the interval of co-occurrence of $w_r$ and $z_s$ shares a common interval with the common interval of the occurrences of both $x_p$ and $y_q$. The algorithm systematically reduces the number of $w_r$ and $z_s$ pairs to be explored in the process of finding the existence of the coincidence.

## 1. Introduction

The problem of coincidence of eventualities[1] within the context of temporal double recurrence has been introduced and studied in [1]. Let x and y be sequences of eventualities [6] such that x is the eventuality sequence $x_0, x_1, \ldots, x_{n-1}$ and y is the eventuality sequence $y_0, y_1, \ldots, y_{m-1}$ such that each $x_i$ and $y_j$ are eventualities and each eventuality has a fixed duration in all its occurrences. An eventuality is deemed to have occurred over a time interval, if an incidence of it takes place exactly over the temporal extents of that time interval. When a sequence x occurs over an interval k, it means k is divided into a sequence of intervals $k_0$, $k_1$, $k_2$,..,$k_{n-1}$ such that $x_i$ occupies $k_i$ for all i such that $0 \leq i \leq n-1$ and $k_i$ meets $k_{i+1}$ (*a la* Allen[3]) for all i such that $0 \leq i \leq n-2$.

[1] Eventualities according to Galton[5] is a generic term that includes events, states, processes, actions or such entities as may have temporal extents. In other words, these are "things" we may wish to put on our calendar.

If an eventuality sequence x recurs over a time interval k, then k can be divided into subintervals $k_0$, $k_2$,… $k_{n-1}$, $n \geq 1$, such that $k_i$ *meets* $k_{i+1}$ (*a la* Allen[3]) for all i such that $1 \leq i \leq n-1$ and a distinct incidence of x occurs over each interval $k_i$.. This captures the notion as defined in the literature such as [2, 6]. If we have the eventuality sequence x recurring over an interval, we can use the sequence $x^1$, $x^2$,….., $x^j$ to refer to intervals over which distinct incidences of x occur and $x^i$ meets $x^{i+1}$ for all i such that $1 \leq i < j$ and j is the number of distinct incidences of x that occur over the interval k. For the $m^{th}$ eventuality in the sequence x, i.e. $x_m$, the term, $x_m^i$ refers to the $i^{th}$ occurrence of $x_m$ within the interval k over which the eventuality sequence x recurs. Let len(x) denote the length of the eventuality sequence x, which is also the number of individual eventuality in x.

If both of the eventuality sequences x and y recur over an interval k, a double recurrence, then the coincidence problem is to determine the existence or otherwise of two non-disjoint subintervals of k (which share a common subinterval) occupied each by incidences of $x_p$ and $y_q$. In other words, determine whether there exists i and j such that $x_p^i$ and $y_q^j$ which are subintervals of k and are non-disjoint. *If this is so, then a coincidence of $x_p$ and $y_q$ is said to exist during k.* The coincidence of $x_p$ and $y_q$ within a double recurrence may be desirable for a "liveness" condition, which is the completion of a plan, to be accomplished or may be undesirable for a safety condition to be maintained. If eventuality sequences x and y both recur over the same interval k, a cycle of such a double recurrence is, roughly speaking, a minimal time subinterval j of k such that each of $x_0^1$ and $y_0^1$ start j and j is finished by both the intervals $x_{n-1}^i$ and $y_{m-1}^j$, where n and m are lengths of eventuality sequences x and y respectively. It has been argued in [1] that the duration of a cycle of a double recurrence is the *lowest common multiple* of the durations of the eventuality sequences x and y.

A simple example of the coincidence problem within the context of a temporal double recurrence (from [1]) is presented as follows. A factory is in a continuous recurrence between a working state and a resting state. The working state takes 5 days while a resting state takes 3 days. Regular maintenance must take place during the resting days. However, the maintenance engineer will not be available on a Wednesday. Thus there are two recurrences here. The first is a recurrence of weekdays: Monday, Tuesday…..Sunday, Monday…. (ad infinitum). The duration of each of the eventuality is one day. The second recurrence is that of Working, Resting, Working, Resting, Working… (ad infinitum). Note that while the first recurrence is natural, the second is artificial according to the classification of recurrences by Carriero et al[4]. The coincidence of interest here then, is that between the resting state and the weekday Wednesday. In the case of this example, we seek for maintenance to be possible. For that to happen, the maintenance engineer must be available for three consecutive days.

It should be noted that the notion of recurrence of eventuality sequences as discussed in this paper is similar to the notion of recurrence of eventuality sequences in the work of Koomen[6] and the eventualities involved exhibit the properties of repetition, periodicity and composition as described by Terenziani and Anselma[8]. What we mean by the recurrence of an eventuality sequence is the consecutive repetition of sequence in such a way that there is no time gap between the two consecutive occurrences. This should be contrasted with notion of recurrence of the recurrence of an eventuality discussed from an ontological standpoint by Carriero et al [4], which captures the idea that an eventuality (or situation) recurs at regular periods, which in a sense is equivalent to the notion of Terenziani and Anselma's periodic events[8].

Eventuality sequences are composite by nature because they consist of one or more individual eventualities. Each of the eventualities in the sequences is periodic and is in fact *strongly periodic* in the Terenziani[9] sense, such that a new instance of the eventuality holds at regular intervals. A strongly periodic eventuality contrasts with a *nearly periodic* eventuality, which is an eventuality that is repeated during every demarcated period say within any week, but not at regular intervals. An example of a strongly periodic eventuality is a meeting that holds every Thursday at 10:00am, while an example of a nearly periodic eventuality is a meeting that holds once a week but can take place on any week day or at any time.

While the definition of a strongly periodic eventuality and Carriero et al's recurrent situations talk about eventualities whose instances takes place at equal intervals, it is unclear about the exactness of their durations. This is because, the definition holds that the duration of the interval between the start of any two succeeding instances of such an eventuality is fixed, while the definition is not specific about whether or not the duration of the time intervals over which the instances of the eventuality hold is fixed. For example, a doctor's visit every Tuesday at 8:00 am is a strongly periodic. However the duration of the visit may differ from week to week. Thus, it is possible to go beyond the definition of strong periodicity to define *exact periodicity*. An eventuality is said to be *exactly periodic* if and only if apart from being strongly periodic, the duration of the time intervals over which any instance of the eventuality hold is the same. It is important to note that each of the eventualities in the recurring sequences we consider in this paper, have an exact periodicity.

One area in which recurrent eventualities naturally occur is in planning and plan reasoning. Koomen[6] for example, demonstrated the need to integrate recurrence reasoning within a planning mechanism for the purpose of allowing a planner to reason about the fulfillment of an action's precondition which is an eventuality within a recurrence that is external to the planning process. For example Koomen gave the example of a robot driver with a plan to drive from one point A to another point B. At some junction point C between the starting point and the destination, there is a traffic light which switches between red and green every 2 minutes. Thus the precondition for the action of moving from C to the destination to commence is for the traffic light to be green. Koomen[6] devised a reified logic that is based on Allen's interval logic for reasoning with recurrence which enables the robot driver to reason that if on the completion the action of moving from the starting point A, to C, it encounters a red light at the junction, it can wait for the precondition it needs to commence the next action in the plan. In order to motivate the need to consider a situation that requires reasoning about coincidence within the context of double recurrence, let there be a gate at C that is shut for 15 minutes at the top of every hour in order to give priority to some tasks of higher priority using the same road. Then the precondition for carrying out the task of driving from C to the destination B, becomes the conjunction of two conditions which are: for the traffic light to be green and for the gate at C to be open, a liveness condition. It is clear that each of the conditions that make up the conjunction is a part of one of the given recurrences that constitute a double recurrence.

Similarly, Terenziani and Anselma[8] encountered recurrence in plan reasoning. Their task was to monitor the implementation of plans for treating medical patients. Such plans usually contain composite and periodic events. Besides, each of the events involved are exactly periodic as described earlier. Their goal is to use the plan to determine whether or not the actions taken conform to the plan. This requires reasoning about temporal constraint satisfaction in order that temporal relations between the intervals over which different actions take place conform to the relations between those of the corresponding action types prescribed in the treatment plans.

However, in addition to this, a different kind of plan reasoning will be required if the patient to be treated suffers from two different pathological conditions that require for him to undergo two parallel treatment plans; one for each of those conditions and for some considerable length of time. Suppose also that within the two sequences of eventualities that make up the plans, there is a pair of eventualities, one from each of the treatment plans, which must not happen around the same time in order, for example, to avert a potentially life threatening drug interaction. This kind of plan reasoning is different from the plan monitoring carried out by Terenziani and Anselma[8]. This requires determining whether or not coincidence occurs between one eventuality from one plan and another eventuality from another, as a result of commencing the two parallel plans at the same time. That is another example of having to reasoning about coincidence within the framework of a double recurrence. This is needed to avert the violation of a safety condition.

This paper presents two algorithms for detecting such coincidences of eventualities within the context of temporal double recurrence, as that which may arise between parts of two recurring plans or that which may arise between eventualities with the a double recurrence of two eventuality sequences. The first one built on the expectation that a coincidence should exist within a double recurrence if and only if it exists within a cycle of the double recurrence. That algorithm runs in quadratic time of the durations of the two eventuality sequence. The second algorithm is the worst case linear-time algorithm that solves the same problem. The algorithm is a product of the properties of the notion of gcd-partition of a double recurrence.

The rest of this paper is organized as follows. Section 1.1 presents the basic notations and the model of time used in this paper and then presents the first algorithm for solving the coincidence problem. The notion of gcd partition and its properties are discussed in section 2, while the algorithm itself and its running time analysis are presented in section 3. The paper is summarized in section 4.

### 1.1 Notation, Time Model and a Preliminary Algorithm

The model of time used in this paper is linear (as opposed to branching) and discrete (as opposed to dense). The paper also uses both time intervals and time points or instants. Each time point can be treated as a natural number. There are equality or inequality relations: =, <, >, ≤ and ≥ that may hold between pairs of time points. Each of the relations has the same signature:

$$=, <, >, \leq, \geq : \text{Nat} \times \text{Nat} \rightarrow \text{Boolean}$$

The start and end functions take a time interval as its argument and returns a time point. Their signature is given thus:

$$\text{start, end} : \text{Interval} \rightarrow \text{Nat}$$

A time interval should be regarded as a pair of two interval numbers, in which the second is greater than the first. As for time intervals, there are standard binary intervals among time intervals that include all of Allen's standard relations, which include Overlaps, Starts, Finishes, Meets etc. However in this paper, relations that will be used will include Subinterval and Disjoint(see [2] and [6]) with the standard signature:

$$\text{Subinterval, Disjoint: Interval} \times \text{Interval} \rightarrow \text{Boolean}$$

The Subinterval relation is also denoted as $\sqsubseteq$ and is defined as equivalent to a disjunction of Allen's relations: Starts, Finishes, Contains or Equals. The Disjoint relation on the other hand is equivalent of the disjunction of Allen's Before or After or Meets or Met-by. Both the inverse relation of Disjoint (otherwise known as the non-disjoint relation) and Subinterval relations can be defined in terms of start and end functions for intervals thus:

For all j, k. $\neg$Disjoint(j, k) if and only if

$$(\text{start}(j) \leq \text{start}(k) < \text{end}(j)) \text{ or } (\text{start}(k) \leq \text{start}(j) < \text{end}(k))$$

For all j, k. Subinterval(j, k) if and only if

$$(\text{start}(k) \leq \text{start}(j)) \text{ and } (\text{end}(j) \leq \text{end}(k))$$

The definition for the non-disjoint relation is a disjunct of two range inequalities that are inclusive on one side and exclusive on the other. This is because in Allen's interval relations, two meeting intervals are treated as disjoint. The definition of the non-disjoint relation is used extensively in Algorithm 2 for coincidence in section 3 to determine non-disjointedness.

The common function is a partial function that is only defined when two time intervals are non-disjoint. It takes two time intervals and returns the maximal common subinterval of the two. The signature is:

$$\text{common} : \text{Interval} \times \text{Interval} \rightarrow \text{Interval}$$

It is important to note that:

For all j, k, it is the case that common(j, k) = common(k, j).

There are a number of relations between the common function and the Disjoint and the $\sqsubseteq$ relation. For example, two time intervals have a common subinterval if and only if they do not have a disjoint relationship. This is expressed as Axiom 1.0(a). The common interval of the common intervals of k with each of j and m is a subinterval of the common interval of j and m. This is expressed as Axiom 1.0(b). The existence of the common interval of the common intervals of k with m and that of k with j imply the existence of a common interval between j and m.

**Axiom 1.0**

*(a.)There exists j = common(k, m) if and only if ¬Disjoint(k, m)*
*(b.) common(common(k, j), common(k, m)) ⊑ common(j, m).*
*(c.) There exists n such that n = common(common(k, j), common(k, m)) implies*
*There exists i such that i = common(j, m).*
*(d) common(common(k, j), common(k, m)) = common(k, common(j, m)).*

Before proceeding here there is a need to formally define what it means for two eventualities to coincide within a certain time interval. The notion of coincidence is different from the idea of two eventualities, $x_p$ and $y_q$ co-occurring over the same interval k i.e. Occurs($x_p$, k) and Occurs($y_q$, k). The fact that two eventualities x and y coincide within a time interval k is denoted by the notation Coincidence(x, y, k). The signature of Coincidence is given by:

Coincidence : Eventuality × Eventuality × Interval → Boolean.

*Two eventualities x and y are said to coincide within an interval k if and only if one interval of occurrence exists each of x and y that are both subintervals of k and those two intervals are non-disjoint.*

**Definition 1.1**

*For all eventualities x, y, ω, Coincidence (x, y, k) is true if and only if*
*There exists j and m such that Occurs(x, j) and Occurs(y, m) and ¬ Disjoint(j, m) and both j, m ⊑ k.*

This paper is also dealing with sequences of eventualities. Eventualities are states or events or processes that may take place over time. A sequence of eventualities is also an eventuality. Thus one may regard an eventuality sequence as an eventuality composed by a binary sequence function, *seq*, with the signature:

seq : Eventuality × Eventuality → Eventuality

The eventuality formed by the application of seq to pair of eventualities x and y, is an eventuality whose occurrence is defined thus:

For all eventualities x, y and interval k, Occurs(seq(x, y), k) is true if and only if
There exists intervals j and i such that Occurs(x, j) and Occurs(y, i) and Meets(j, i).

Thus, an eventuality sequence x of length n can be thought of as comprising of the following applications of the seq functions:

seq($x_0$, seq($x_1$, seq($x_2$, seq($x_3$, …seq($x_{n-2}$, $x_{n-1}$)…)))).

Thus, the expression above is a formal expression for the eventuality sequence x. A double recurrence is simply represented as a pair of the recurring eventualities e.g. (x, y). The set of all the intervals that are cycles of the double recurrence of a double recurrence (x, y) is denoted by Ω(x, y). Thus the fact that an interval ω is a cycle of the double recurrence of (x, y) is denoted by ω∈ Ω(x, y). The formal definition of a cycle is given below:

**Definition 1.2 (A Cycle of Double Recurrence)**

*The interval ω is a cycle of the double recurrence of x and y (i.e. ω ∈ Ω(x, y)) if and only if*

- *Both of eventualities x and y recur over the interval ω.*
- *The interval ω is a minimal interval such that:*
  - *There exist intervals j and k such that Occurs(x, j) and Occurs(y, k) and both j and k start ω.*
  - *There exists intervals m and n such that Occurs(x, m) and Occurs(y, n) and both m and n finish ω.*

Every eventuality has a fixed duration. Thus there is a duration function with the following signature:

duration : Eventuality → Nat

Restricting durations to natural numbers is in keeping with a discrete model of time. As such, the minimum value of any duration is 1, because there are no zero-duration intervals. Besides we eliminate the need for duration values that are real values, on the account of the fact that if there are any durations values given to n decimal places, the unit of duration can be multiplied by $10^n$ in order to make it and other duration values with less than n decimal places, integers.

On the basis of a result from the literature [1] which states that each cycle of recurrence is an exact copy of others, (in other words, the temporal relationship between the $i^{th}$ $x_p$ and the $j^{th}$ $y_q$ within any cycle of the double recurrence of x and y is an invariant), it follows that a coincidence occurs in one cycle if and only if it exists in all the others. This is a direct consequence of the fact that the duration of any eventuality is fixed from occurrence to occurrence. Therefore an algorithm exists for the coincidence problem which only needs to explore an arbitrary cycle of the double recurrence looking for a coincidence. If that fails to find a coincidence, then no coincidence can ever occur between $x_p$ and $y_q$ during the entire interval over which a double recurrence of x and y happens. Otherwise, then a coincidence is guaranteed. That algorithm is formally presented here as Algorithm 1. The running time of that algorithm is O(duration(x)*duration(y)) in the worst case.

<u>Algorithm 1: Temporal Projection Algorithm for Coincidence</u>

*projection-coincidence (x, y : eventuality_sequence, p: index of x, q: index of y) returns Boolean*
*/* find out whether $x_p$ and $y_q$ coincide */*
*begin*
*r ← 0*
*s ← 0*
*sum1 ← 0*
*sum2 ← 0*
*reset flag*
*repeat*
        *begin*

```
            if r = p and s = q then return true else skip
            if sum1 + duration(x_r) = sum2 + duration(y_s) then
                  if r = len(x) - 1 and s = len(y) - 1 then set flag
                  else    begin
                                sum1 ← sum1 + duration(x_r)
                                sum2 ← sum2 + duration(y_s)
                                inc r mod len(x)
                                 inc s mod len(y)
                          end

            else  if sum1 + duration(x_r)  >  sum2 + duration(y_s) then
                          begin
                                sum2 ← sum2 + duration(y_s)
                                increment s mod len(y)
                          end
                  else    begin
                                sum1 ← sum1 + duration(x_p)
                                increment r mod len(x)
                          end
      end
until flag
return false
end
```

The next theorem formally expresses the result about the running time of the Algorithm, while the proof of Theorem 1.3 is presented in the Appendix.

**Theorem 1.3**
*The worst case running time of Algorithm 1 is O(duration(x)*duration(y)).*

A partition relationship exists between a pair of eventuality sequences if the latter sequence is deemed to take place over an interval over which the former takes place over the same interval. That relationship is denoted by the binary predicate, Part (formally defined as Definition 2.0) so that Part(x, z) means z is a partition of x. The Part relationship is symmetric. The fact that the gcd-partition relationship exists between a pair of double recurrences, (x, y) and (w, z) is denoted by GCD-Part((x, y), (w, z)) (formally defined in Definition 2.1). It holds when it is the case that Part(x, w) and Part(y, z) and the duration of each of the eventualities in w and z are the same and their duration is the greatest common divisor of the durations of x and y. The GCD-Part relationship is not symmetric.

Because of the partition relationships, the intervals of occurrences of ordered pairs of eventualities from x and w, which are partitions for each other, have definite qualitative relationships within any interval over which x and occur. A pair of individual eventualities have a natural qualitative (temporal) relationship (such as Allen's relations as well as Disjoint and Subinterval)  if every occurrence of one of them implies an occurrence of the other eventuality and their intervals of occurrence satisfy the qualitative relationship. For example, the fact that two individual eventualities are naturally non-disjoint is denoted by NN-Disjoint($x_p$, $y_q$) and is formally defined as Definition 2.2(a), while the formal definitions of natural subinterval and naturally overlap relation among a pair of eventualities are denoted by N-Subinterval, N-Overlaps respectively and is formally defined as Definition 2.2(b) and Definition 2.2(c) respectively.

Section 2 presents the theory of gcd partitions and shows how it is deployed to solve the problem of coincidence.

## 2. GCD Partitions and Properties

This section briefly introduces the notion of gcd partitions for a double recurrence and their formal properties. This notion is very crucial for the understanding of the algorithm to be presented in the next section. We will start by defining the notion of *partitions* in Definition 2.0.

**Definition 2.0**
*An eventuality sequence w is a partition of another eventuality sequence x denoted Part(x, w) if and only if:*

- *Both eventualities x and w have the same duration i.e. dur(x) = dur(w)*
- *For any interval k, Occurs(x, k) if and only if Occurs(w, k).*

Building on the definition of partitions we will define gcd-partitions for a double recurrence in Definition 2.1.

**Definition 2.1**
*A double recurrence (w, z) is a gcd partition of another double recurrence (x, y) denoted as GCD-Part((x, y), (w, z)) if and only if*

- *Part(x, w)) and Part(y, z)*
- *For all eventuality* $x_p$ *and* $y_q$*, such that* $0 \leq p \leq len(x)$ *and* $0 \leq q \leq len(y)$*, it is the case that* $dur(x_p) = dur(y_q)$ *= gcd of dur(x) and dur(y).*

When an eventuality sequence is defined as the partition of another eventuality sequence, for example say w is a partition of x, then, there is a fixed set of temporal relations between the interval of occurrence of any eventuality in x and that of any (say $x_p$) and any eventuality in w (say $w_r$) that exist within any interval of occurrence of x.

**Definition 2.2**
*If for two eventuality sequences x and w, it is the case that Part(x, w), then for any p and r such that* $0 \leq p \leq len(x)$ *and* $0 \leq r \leq len(w)$*, then*

(a) *Eventualities* $x_p$ *and* $w_r$ *are said to be naturally non-disjoint if and only if every interval of occurrence of* $x_p$ *within any interval of x, is non-disjoint with some interval of occurrence of* $w_r$*, within the same interval of occurrence of x. It is denoted as NN-Disjoint(*$w_r$*,* $x_p$*).*

(b) *Eventuality $w_r$ is said to be a natural subinterval of eventuality $x_p$, if and only if every interval of occurrence of $w_r$ is a subinterval of some interval of occurrence of $x_p$. The relation is denoted as N-Subinterval($w_r$, $x_p$).*

(c) *Eventuality $w_r$ is said to naturally overlaps eventuality $x_p$, if and only if every interval of occurrence of $w_r$ overlaps some interval of occurrence of $x_p$. The relation is denoted as N-Overlaps($w_r$, $x_p$).*

It is always the case that if w is a partition of x, then any eventuality in x, say $x_p$, there is an eventuality in w, with which it is non-disjoint. That is described in Axiom 2.3.

**Axiom 2.3**

*If w is a partition of x, then, there exists an occurrence of some eventuality in w that is non-disjoint with any subinterval of the occurrence of any eventuality in x, i.e.*

*For all x, w, p, k. if Part(x, w) and Occurs($x_p$, k) for $0 \leq p \leq len(x) - 1$ then*

*for any $k_1$ such that $k_1 \sqsubseteq k$ there exists r such that $0 \leq r \leq len(w)-1$ and intervals j such that: Occurs($w_r$ , j) and ¬Disjoint($k_1$, j).*

Theorem 2.4 describes the fact that the relative distance between the starts and the ends of the intervals of occurrence of any $x_p$ and that of $w_r$ for all possible values of p and r, is invariant within every interval of occurrence of x.

**Theorem 2.4**

*Let the eventuality sequence w be a partition of eventuality sequence x. For any pair of eventualities $x_p$ and $w_r$ from x and w respectively, and let $x_p$ and $w_r$ be intervals over which they both occurred within a specific interval of occurrence of x, then start($x_p$) – start($w_r$) and end($x_p$) – end($w_r$) are invariant in different intervals of occurrence of x.*

It should be noted that Theorem 2.4 here is a consequence of the fact that all occurrences of each eventuality have a fixed duration. The following definition formally introduces a parallel operator + for the co-occurrence of two eventualities.

**Definition 2.5 (Parallel Operator + for Co-occurrence)**

*An eventuality denoted by $x_r + y_s$ is one defined by the co-occurrence of eventualities $x_r$ and $y_s$, so that for any interval j, stating that Occurs($x_r + y_s$, j)is equivalent to stating that: Occurs($x_r$, j) and Occurs($y_s$, j).*

The following theorem, Theorem 2.6, is a basic mathematical consequence of the nature of gcd partitions.

**Theorem 2.6**

*If (w, z) is a gcd-partition for any sequence of eventuality pair (x, y), the lengths of w and z, denoted as len(w) and len(z), are both prime numbers.*

The following theorem (Theorem 2.7) is the key property of gcd-partitions that will facilitate the construction of our proposed algorithm. The proofs of Theorems 2.7, 2.8, 2.9, 2.10, 2.11 and 2.12 are presented in the Appendix.

**Theorem 2.7**

*If the eventuality sequence pair (w, z) is a gcd-partition of the eventuality sequence pair (x, y), then for any cycle k of the double recurrence of (w, z) and for any r and s such that* $1 \leq r \leq len(w)$ *and* $1 \leq s \leq len(z)$*, there exists a subinterval of k over which the occurrences of both* $w_r$ *and* $z_s$ *happens.*

The proof of Theorem 2.7 is contained in the Appendix. It should be noted that following from Theorem 2.7, any cycle of the double recurrence (x, y) can be divided into a sequence of consecutive (or meeting) intervals. The duration of each interval is the greatest common divisor of the durations of x and y, and for each of these intervals, i, there exists r and s, within the bounds of the number of eventualities in the eventuality sequences w and z that constitute the gcd-partition of the double recurrence (x, y), such that it is the case that: Occurs($w_r$, i) and Occurs($z_s$, i).

A corollary that follows from Theorem 2.7 expresses the fact that any eventuality sequence that holds over the same interval as the cycle of a double recurrence of an eventuality sequence has a partition of another eventuality sequence defined as a sequence of co-occurrences of eventualities $w_0 + z_0$, $w_1 + z_1 \ldots w_{len(w)-1} + z_{len(z)-1}$ from the gcd-partition (w, z) of the double recurrence.

**Corollary 2.7a**

*Let GCD-Part((x, y), (w, z)) then for any interval* $\omega \in \Omega(x, y)$ *then it is the case that for any eventuality sequence u that is deemed to occur over any cycle of the double recurrence (x,y) (i.e. For all u and* $\varphi$ *such that Occurs(u,* $\varphi$*) if and only if* $\varphi \in \Omega(x, y)$*) then there exist another eventuality sequence v such that Part(u, v) and len(v) = len(w) * len(z) and*

$$v_0 \equiv w_0 + z_0,$$
$$v_{r+1} \equiv w_{index(w,\, r)+1 \bmod len(w)} + z_{index(z,\, r)+1 \bmod len(z)}, \text{ for all } 0 \leq r < len(v) - 1$$
$$v_{len(v)-1} \equiv w_{len(w)-1} + z_{len(z)-1}$$

*where index(w, r)means the index of w in the definition of* $v_r$*.*

Therefore, any subinterval of k must share a common subinterval with at least one of intervals arising from that division. That is formally stated as Lemma 2.7b.

**Lemma 2.7b**

*Let the interval* $\omega$ *be such that:* $\omega \in \Omega(x, y)$*, and GCD-Part((x, y), (w, z)), then, for any subinterval k of* $\omega$*, (i.e.* $k \sqsubseteq \omega$*) there exists another subinterval m of* $\omega$ *(i.e.* $m \sqsubseteq \omega$*) such that there exists r, s for which Occurs(*$w_r + z_s$*, m) and it is the case that m and k are non-disjoint (i.e.* $\neg$*Disjoint(k, m)).*

**Theorem 2.8**

*If the pair of eventuality sequence (w, z) is a gcd-partition for the double recurrence eventuality sequence pair (x, y) and there exists an interval $\omega$ such that $\omega \in \Omega(x, y)$.*

*Let there exist r, s such that $0 \leq r \leq len(w) \wedge 0 \leq s \leq len(z)$ such that*
*NN-Disjoint($x_p$, $w_r$) and NN-Disjoint($y_q$, $z_s$) then*

*Let there be intervals k, $k_1$, j, m such that k, $k_1 \sqsubseteq \omega$ and Occurs($w_r$, k) and Occurs($z_s$, $k_1$), and that Occurs($x_p$, j) and Occurs($y_q$, m) it is the case that (where k and j are within the same interval of occurrence of x and $k_1$ and m are within the same interval of occurrence of m) then*

*A coincidence exists between $x_p$ and $y_q$ within $\omega$ for $0 \leq p \leq len(x) -1$ and $0 \leq q \leq len(y) -1$*
*if and only if*
*$k = k_1$ implies that $\neg$Disjoint(common(k, j), common($k_1$, m)).*

The next four theorems (Theorem 2.9-2.12) introduce particular instances of Theorem 2.8 in which the pair of temporal relationships between $x_p$ and $y_q$ and the specific eventualities from the partitions of x and y which constitute the gcd partition of the double recurrence (x, y) (i.e. (w, z)), with which it is non-disjoint is specified. In each case the specified relations guarantees that a coincidence will happen within a cycle of double recurrence.

Each of these theorems considers specific instances of non–disjoint relations between $x_p$ and $w_r$ on one hand and $y_q$ and $z_s$ on the other. From an algorithmic standpoint, each theorem will consider the number of $w_r$ and $z_s$ pairs with which $x_p$ and $y_q$ are naturally disjoint in determining coincidence.

Theorem 2.9 states that coincidence is guaranteed once an eventuality from w that is a natural subinterval of $x_p$ or there exists an eventuality from z that is a natural subinterval of $y_q$.

**Theorem 2.9**
*If the pair of eventuality sequence (w, z) is a gcd-partition for an eventuality sequence pair (x, y), then:*
*A coincidence exists between intervals of occurrence of $x_p$ and $y_q$ within a cycle, $\omega$ of the double recurrence of $\omega$, (denoted as Coincidence($x_p$, $y_q$, $\omega$)) for $0 \leq p \leq len(x)-1$ and $0 \leq q \leq len(y)-1$ within any interval $\omega$ such $\omega \in \Omega(x, y)$ if*

- *There exists: r, s. $0 \leq r \leq len(w)$ and $0 \leq s \leq len(z)$ such that*
  *N-Subinterval($w_r$, $x_p$) or N-Subinterval($z_s$, $y_q$).*

The condition for determining coincidence in Theorem 2.9 is for either of $w_r$ or $z_s$ to be a natural subinterval of either of $x_p$ or $y_q$ respectively. By testing the condition of Theorem 2.9 first in algorithm 2, the algorithm reports coincidence in the case for which either $x_p$ or $y_q$ is naturally disjoint with more than two eventualities from the partitioning sequences, w and z, respectively.

Thus, checking this condition first, eliminates the need to subsequently explore more than four pairs of eventualities from the gcd-partitions w and z, with which either or both of $x_p$ and $y_q$ are naturally disjoint in exploiting Theorem 2.8.

Theorem 2.10 states that coincidence is guaranteed once each of $x_p$ and $y_q$ naturally overlaps an eventuality from w and z respectively. This is the only possibility that the algorithm would have had to consider four pairs of eventualities from w and z in the course of deciding the existence of coincidence. However, in this case, coincidence is guaranteed.

**Theorem 2.10**
*If the pair of eventuality sequence (w, z) is a gcd-partition for an eventuality sequence pair (x, y), then:*
*A coincidence exists between intervals of occurrence of $x_p$ and $y_q$ (denoted as Coincidence($x_p$, $y_q$, $\omega$)) for $0 \leq p \leq len(x)-1$ and $0 \leq q \leq len(y)-1$ within any interval $\omega$ such that $\omega \in \Omega(x, y)$ if*

- *$\exists r, s.\ 0 \leq r \leq len(w)-2 \wedge 0 \leq s \leq len(z)-2$ such that*
  *N-Overlaps($w_r$, $x_p$) and N-Overlaps($z_s$, $y_q$).*

After checking the condition of Theorem 2.9 in the algorithm, the next condition to be checked is that of Theorem 2.10 above. Consequently, there are two other remaining possibilities:

- The case in which $x_p$ ($y_q$) naturally overlaps an eventuality in w (z), and $y_q$ ($x_p$) is a natural subinterval of an eventuality in z (w). (The case addressed by Theorem 2.12).
- The case in which $x_p$ is a natural subinterval of an eventuality in w and $y_q$ is a natural subinterval of an eventuality in z. (The case addressed by Theorem 2.11).

**Theorem 2.11**
*If the pair of eventuality sequence (w, z) is a gcd-partition for an eventuality sequence pair (x, y) and*

*There exists r and s such that $0 \leq r \leq len(w)-1 \wedge 0 \leq s \leq len(z) -1$ and N-Subinterval($x_p$, $w_r$) and N-Subinterval($y_q$, $z_s$) then*

*A coincidence exists between intervals of occurrence of $x_p$ and $y_q$ within any interval $\omega \in \Omega(x, y)$ (denoted as Coincidence($x_p$, $y_q$, $\omega$) for $0 \leq p \leq len(x) - 1$ and $0 \leq q \leq len(y) - 1$ within $\omega$ if and only if*

> *For any intervals j, k, $j_1$, $k_1 \sqsubseteq \omega$ and $\omega$, such that Occurs($w_r$, j) and any interval k of occurrence of $x_p$ which is not disjoint with j (within an interval of occurrence of x) and for any interval $j_1$ such that Occurs($z_s$, $j_1$) and any interval $k_1$ of occurrence of $y_q$ which is not disjoint with $j_1$*
>
> $$start(k) - start(j) \leq start(k_1) - start(j_1) < end(k) - start(j) \text{ or}$$
> $$start(k_1) - start(j_1) \leq start(k) - start(j) < end(k_1) - start(j_1).$$

The final theorem, Theorem 2.12 is here presented.

**Theorem 2.12**
*If the pair of eventuality sequence (w, z) is a gcd-partition for an eventuality sequence pair (x, y) and*

*There exists r and s such that* $0 \leq r \leq len(w)-1 \wedge 0 \leq s \leq len(z)-2$ *and*

*N-Subinterval*($x_p$, $w_r$) *and N-Overlaps*($z_s$, $y_q$) *and N-Overlaps*($y_q$, $z_{s+1}$) *then*

*A coincidence exists between intervals of occurrence of* $x_p$ *and* $y_q$ *(denoted as Coincidence*$\omega$($x_p$, $y_q$)) *for* $0 \leq p \leq len(x) -1$ *and* $0 \leq q \leq len(y) -1$ *within* $\omega$ *if and only if*

> *For any intervals* $j, k, j_1, k_1 \sqsubseteq \omega$ *and* $\omega$*, such that Occurs*($w_r$, $j$) *and any interval k of occurrence of* $x_p$ *(where j and k are within an interval of occurrence of x) and for any interval* $j_1$ *such that Occurs*($z_s$, $j_1$) *and any interval* $k_1$ *of occurrence of* $y_q$ *(where* $j_1$ *and* $k_1$ *Sare within the same occurrence of y).*
>
> $start(k) - start(j) \leq start(k_1) - start(j_1) < end(k) - start(j)$ *or*
>
> $start(k_1) - start(j_1) \leq start(k) - start(j) < end(j_1) - start(j_1)$ *OR*
>
> $0 \leq start(k) - start(j) < end(k_1) - end(j_1)$ .

The gcd partition algorithm is presented in the next section. The algorithm will take advantage of the theorems presented in this section.

## 3. The GCD Partition Algorithm

The algorithm works by identifying the common subintervals that each of $x_p$ and $y_q$ share with the eventualities from the sequences w and z (which together constitute the gcd-partition of double recurrence (x, y)) respectively, and then deciding whether or not the intervals of occurrences of $x_p$ and $y_q$ will coincide as a result of the fact that some eventuality pair ($w_r$, $z_s$) from w and z share the same exact interval of occurrence following Theorem 2.8. The coincidence of $x_p$ and $y_q$ is guaranteed within any cycle of the double recurrence of eventuality sequences x and y if and only if such a pair ($w_r$, $z_s$) exists which guarantees that $x_p$ and $y_q$ share a common subinterval with the interval of co-occurrence of both $w_r$ and $z_s$.

Algorithm 2: The gcd-partition Algorithm

1. *gcd-partition-coincidence(x, y : eventuality_sequence p, q : index) returns Boolean*
2. *begin*
3. *g = gcd(duration(x), duration(y))*
4. *prexp* $\leftarrow$ *start of eventuality* $x_p$ *(i.e.* $\sum_{k=0}^{p-1} duration(x_k)$)
5. *preyq* $\leftarrow$ *start of eventuality* $y_q$ *(i.e.* $\sum_{k=0}^{q-1} duration(y_k)$)
6. *postxp* $\leftarrow$ *prexp + duration*($x_p$) /* *end point of* $x_p$*/
7. *postyq* $\leftarrow$ *preyq + duration*($y_q$) /* *end point of* $y_q$ */
8. /**identify the index of the first gcd eventuality,* $w_r$, *with which* $x_p$ *is non-disjoint* */
9. $r \leftarrow \lfloor prexp/g \rfloor$
10. /* *identify the index s of the first gcd eventuality,* $z_s$, *with which* $y_q$ *is non-disjoint*/
11. $s \leftarrow \lfloor preyq/g \rfloor$
12. /* *Theorem 2.9: if* $w_r$ *or its successor is a subinterval of* $x_p$ *or* $z_s$ *equals or is within*
13. /* $y_q$ *then true* */
14. if (prexp = r*g and postxp >= (r+1)*g) or( prexp < (r+1)*g and postxp >(r+2)*g)
15. *then* *return true*

16. else /* Th. 2.9: if $z_s$ or successor is a subinterval of $y_q$ or $z_s$ is a subinterval of $y_q$*/
17. if (preyq = s*g and postyq >= (s+1)*g) or (preyq < (s+1)*g and
18. postyq >(s+2)*g) then return true
19. else skip
20. startxp1 ← beginning of the common interval of $x_p$ and $w_r$ relative to start of $w_r$
21. endxp1 ← end of common interval of $x_p$ and $w_r$ relative to start of $w_r$
22. startxp2 ← beginning of common interval of $x_p$ and $w_{r+1}$ relative to start of $w_{r+1}$
23. endxp2 ← the end of common interval of $x_p$ and $w_{r+1}$ relative to start of $w_{r+1}$
24. startyq1 ← beginning of common interval of $y_q$ and $z_s$ relative to start of $z_s$
25. endyq1 ← end of the common interval of $y_q$ and $z_s$ relative to start of $z_s$
26. startyq2 ← beginning of the common interval of $y_q$ and $z_{s+1}$, relative to start of $z_{s+1}$
27. endyq2 ← the end of the common interval of $y_q$ and $z_{s+1}$ relative to $z_{s+1}$.
28. /* Lemma 2.7: if the occurrence of $x_p$ extends beyond $w_r$ and the occurrence y extends*/
29. /* beyond $z_s$ */
30. if startxp2 =0 and startyq2 = 0 then return true /* Theorem 2.10*/
31. else if startxp2 and startyq2 are both -1 then /* Theorem 2.11 */
32. return startxp1 ∈ [startyq1, endyq1) or startyq1 ∈ [startxp1, endxp1)
33. else
34. if startxp2 = 0 and startyq2 = -1 then /* Theorem 2.12 */
35. return startxp1 ∈ [startyq1, endyq1) or
36. startyq1 ∈ [startxp1, endxp1) or staryq1 ∈ [startxp2, endxp2)
37. else if startxp2 =-1 and startyq2 =0 then /* Theorem 2.12 */
38. return startyq1 ∈ [startxp1, endxp1) or
39. startxp1 ∈ [startyq1, endyq1) or startxp1 ∈[startyq2, endyq2)
40. else skip
41. end

The details of each case depend on the nature of the natural relationships between $x_p$ and naturally non-disjoint eventualities from w, and that between $y_q$ and naturally non-disjoint eventualities from z. If the occurrence of either of $x_p$ equals or has within it, the occurrence of an eventuality from w (i.e. if an eventuality from w is a natural subinterval of $x_p$) or $y_q$ has a similar relationship with any eventuality from the sequence z, then there is a guarantee according to Theorem 2.9, that there exists intervals of occurrences of $x_p$ and $y_q$ within any cycle of the double recurrence of x and y that are non-disjoint. So that coincidence of $x_p$ and $y_q$ holds within any cycle of the double recurrence. In order to eliminate the need for exploring several pairs ($w_r$, $z_s$) that are naturally non-disjoint with $x_p$ and $y_q$ respectively, this condition described here should be the first one to be explored in order to determine coincidence as done in line 14. This is because once the pre-condition for Theorem 2.9 is dealt with and acted upon there will be no need to subsequently consider a situation in which either of $x_p$ or $y_q$ shares a common subinterval with more than two of their respective gcd eventualities. Once the precondition for Theorem 2.9 has been ruled out, the possibility of either $x_p$ or $y_q$ being naturally non-disjoint with more than two consecutive eventualities within the eventuality sequences w or z has been ruled out.

Thus, from this point we assume that if $x_p$ and $y_q$ has yet to be adjudged to be coincidental, then both $x_p$ and $y_q$ cannot be naturally non-disjoint with more than two eventualities from w and z sequences respectively.

The starting and ending points of the common subinterval of the occurrences of $x_p$ relative to the start of a naturally non-disjoint eventuality $w_r$ (and $w_{r+1}$ if it exists) and that of $y_q$ relative to the start of a naturally non-disjoint eventuality $z_s$ (and $z_{s+1}$ if it exists) are then computed in lines 20-27 of the algorithm. The variables startxp1 and endxp1 denote the starting and the ending points, respectively, of the common subinterval of the interval of occurrence of $x_p$ and the interval of the first eventuality $w_r$ with which it is naturally non-disjoint, relative to the start of the interval of occurrence of $w_r$. The variables startxp2 and endxp2, also denote the starting and ending points respectively of $x_p$ of the common interval of the occurrence of $x_p$ and that of $w_{r+1}$, (if any) relative to the start of $w_{r+1}$.

Similar definitions exist for startyq1, endyq1 and startyq2 and endyq2. The variables startyq1 and endyq1 denote the starting and ending points respectively of the common subinterval of the intervals of occurrences of $y_q$ and that of the first naturally non-disjoint $z_s$ relative to the start of the interval of occurrence of $z_s$. Similarly, startyq2 and endyq2 denote the starting and ending points of the common subinterval of intervals of occurrences of $x_p$ and that of $z_{s+1}$ (if any) relative to the start of the interval of occurrence of $z_{s+1}$.

It is important to note that startxp2 and startyq2 can either be 0 or -1. The fact that startyq2 = -1, is an indication that the occurrence of $y_q$ is only naturally non-disjoint with the occurrence of the gcd partition eventuality $z_s$ and it is not naturally non-disjoint with the succeeding partition eventuality $z_{s+1}$. On the other hand, startyq2 = 0 means that the occurrence of $y_q$ shares a common subinterval with $w_{r+1}$. All these variables are computed in lines 20 to 27.

If both of startxp2 and startyq2 are both 0, then it means that $x_p$ is naturally non-disjoint with both $w_r$ and $w_{r+1}$, while $y_q$ is also naturally non-disjoint with both of $z_s$ and $z_{s+1}$.Then according to Theorem 2.10, coincidence between a pair of occurrences of $x_p$ and $y_q$ is guaranteed within any cycle of the recurrence of x and y. That condition is checked in line 29.

The next condition to be checked in line 30 is if each of $x_p$ and $y_q$ is a natural subinterval of only $w_r$ and $z_s$. This is a case in which $x_p$ and $y_q$ are naturally disjoint with only one eventuality each from the partitions of x and y respectively. In that case, line 31 returns the result of a Boolean expression which evaluates to true when a coincidence between the occurrence of $x_p$ and $y_q$ is expected to happen as a result the co-occurrence of $w_r$ and $z_s$ over the same interval. That Boolean expression is similar to that presented in definition of the non-disjoint relation between intervals in section 1.1.

The last two conditions to be taken care of are the conditions startxp2 = -1 (while startyq2 = 0) and startyq2 = -1(while startxp2 = 0). With the former condition, every occurrence of $x_p$ is naturally non-disjoint with only one of x's gcd partition eventualities (say $w_r$) while every occurrence of $y_q$ is naturally non-disjoint with two of y's gcd partition eventualities, (say $z_s$ and $z_{s+1}$). In that case, within any cycle of the double recurrence of x and y, the condition for a coincidence to hold for eventualities $x_p$ and $y_q$ under this circumstance is derived from the same disjunction of inequalities in Theorem 2.12 (and it is the Boolean condition returned in lines 38-39 of Algorithm 2) given as follows:

either startxp1 is in between startyq1(inclusive) and endyq1 Or
startyq1 is in between startxp1(inclusive) and endxp1 Or
startxp1 is in between startyq2(inclusive) and endyq2.

Similarly, for the case in which startyq2 = -1, every occurrence of $x_p$ is naturally non disjoint with two of the eventualities in w, ($w_r$ and $w_{r+1}$, say) while $y_q$ is naturally disjoint with only one of y's gcd partition eventualities ($z_s$, say). Thus, within any cycle of the double recurrence of x and y in this circumstance, the condition for a coincidence of the occurrences of $x_p$ and $y_q$ to happen is similar to that given in Theorem 2.12. It is the Boolean condition returned in lines 35-36 and given as follows:

either startxp1 is in between startyq1(inclusive) and endyq1 Or
startyq1 is in between startxp1(inclusive) and endxp1 Or
startyq1 is in between startxp2(inclusive) and endxp2.

Finally, the worst case running time of Algorithm 2 will now be formally shown to be O(max(duration(x), duration(y))). That result is presented as Theorem 3.1.

**Theorem 3.1**
*The running time of Algorithm 2 is O(max(duration(x), duration(y))).*

The proof of Theorem 3.1 is contained in the Appendix. Thus, Algorithm 2 is a more efficient algorithm than Algorithm 1.

## 4. Summary

This paper has presented two algorithms for solving the coincidence problem within the context of double recurrence of two eventuality sequences. The first algorithm presented works by exploring a typical cycle of the double recurrence until it can either conclude the existence of the coincidence of $x_p$ and $y_q$ or the cycle of double recurrence is exhausted without detecting coincidence. The algorithm runs in quadratic time in the duration of the eventuality sequences.

The second algorithm uses the notion of gcd partitions introduced in this paper. The algorithm exploits the key property of gcd partition of a double recurrence (x, y) specified in Theorem 2.7. That property guarantees that within any cycle of double recurrence of x and y, there is a definite interval over which any pair of eventualities ($w_r$, $z_s$) taken from the gcd partition (w, z) of the double recurrence of x and y, co-occur. By Lemma 2.7b, any $x_p$ is naturally disjoint with some an eventuality $w_r + z_s$ formed from an ordered pair of eventualities from w and z. The same is true of the eventuality $y_p$. Thus, if a coincidence must happen between $x_p$ and $y_q$, the common interval of the intervals of occurrence of $x_p$ and $y_q$ must share a common interval with an interval of occurrence of $w_r + z_s$, for some r and s such that $0 \le r < len(w) - 1$ and $0 \le s < len(z) - 1$. That is the essence of Theorem 2.8. The four theorems, Theorem 2.9 to 2.12, are specializations of Theorem 2.8.

By Theorem 2.9, coincidence between $x_p$ and $y_q$ is guaranteed if every interval of occurrence of $x_p$ has a natural subinterval which is an interval of occurrence of any eventuality $w_r$ from w. Again, by Theorem 2.9 coincidence is guaranteed if any interval of occurrence of $y_q$ naturally has a subinterval which is an interval of occurrence of an eventuality $z_s$ from z. By checking the precondition for Theorem 2.9 first, the algorithm eliminates the need to check more than four unique eventualities of the kind $w_r + z_s$ within which interval of occurrence a coincidence of $x_p$ and $y_q$ may happen.

By also checking for the guarantee of coincidence of $x_p$ and $y_q$ as a result of $x_p$ overlapping $w_r$ and $y_q$ overlapping $z_s$, which is the precondition for Theorem 2.10, the algorithm limits the search for coincidence to no more than two pairs of eventualities in w and z.

Algorithm 2 eliminates the need to carry out temporal projection over a full cycle of the double recurrence as described in [1] and presented as Algorithm 1 in this paper. Rather it goes ahead to determine the gcd partitions with which both $x_p$ and $y_q$ are naturally non-disjoint and then applies Theorems 2.9 to Theorem 2.12 to determine the existence of a coincidence of $x_p$ and $y_q$ within any cycle of the double recurrence of x and y. Hence, its worst-case running time is linear, rather than quadratic as that of Algorithm 1 is. Both algorithms have been implemented in Python and tested for correctness with appropriate test data. A more detailed experimental running time analysis of the two algorithms is the subject of ongoing research.

Declaration
The authors have no financial or non-financial interests to disclose.

## Appendix

Proof of Theorem 1.3:

Let the duration of the eventuality sequence x be m and let the duration of the sequence y be n. The length of a cycle of double recurrence is the longest common multiple of m and n. Thus the longest that the length of a cycle can be is m*n. This is the case when the greatest common divisor i.e. gcd(m, n) =1.

The repeat loop in the algorithm explores a single cycle of the double recurrence of x and y and the loop is traversed a maximum of len(x) *len(y) times in order to complete the exploration of cycle. However, since the duration of each eventuality can only be a natural number, the maximum number of times the repeat loop will be explored is when len(x) = duration(x) and len(y) = duration(y). This is when the number of eventualities is the same as the duration. Therefore the worst case running time of the algorithm is duration(x) *duration(y). □

Proof of Theorem 2.7

Let it be the case that len(w) ≥ len(z) and that (w, z) is a gcd-partition of the double recurrence (x, y). Let $w_k^j$ be the $j^{th}$ occurrence of $w_k$ within the double recurrence of w and z. If j =1, then $w_k^1$ occurs over the same interval as $z_k^1$. In general, the occurrence of $w_k^j$ holds over the same interval as an occurrence of $z_{f(j,\, k)}$ where :

$$f(1, k) = k \text{ and } \quad f(j+1, k) = f(j, k) + k \bmod len(z)$$

The function f is a Linear Congruential Generator (LCG) with a maximum period len(z) for the following reasons, according to the properties of LCG[5]:

i. The length of z, is a prime number. Note also that the length of z is a prime number, according to Theorem 2.4. Thus, the only number that divides both len(z) and k is 1.
ii. The only prime number that divides the length of z is itself. But it also divides the coefficient of f(j, k) minus 1 which is 0.
iii. The length of z is not divisible by 4. Thus, there is no need for a -1 to divide 4.

The implication of each incidence of $w_k$ within the cycle of the double recurrence of (w, z) shares exactly the same interval of occurrence with a different eventuality from the z sequence len(z) times. Note also that the number of incidences of w (and ipso facto, $w_k$) within any such cycle is len(z). This also holds for each k with the range 0 ≤ k ≤ len(w)-1. Thus, within a cycle, each of the incidences of $w_k$ shares the same interval of occurrence with a unique eventuality from the sequence z and every eventuality in the sequence z shares some incidence of $w_k$, for all k within 0 and len(w).The proof is concluded. □

Proof of Lemma 2.7b

Let $\omega \in \Omega(x, y)$, then according to Corollary 2.7, for any eventuality u such that Occurs(u, $\varphi$) if and only if $\varphi \in \Omega(x, y)$, there exists another eventuality v which is a partition of u i.e. Part(u, v) and v is defined thus:

$$v_0 \equiv w_0 + z_0,$$
$$v_{r+1} \equiv w_{index(w,\, r)+1 mod\, len(w)} + z_{index(z,\, r)+1 mod\, len(z)},\ \textit{for all } 0 \leq r < len(v) - 1$$
$$v_{len(v)-1} \equiv w_{len(w)-1} + z_{len(z)-1}$$

*where index(w, r)means the index of w in the definition of $v_r$ and len(v) = len(w)*len(z).*

By Axiom 2.3, for any subinterval k of $\omega$, there exists p within bounds $0 \leq p \leq len(v) - 1$, s such that Occurs($v_p$, m) and $\neg$Disjoint(k, m).

The next theorem, Theorem 2.8, shows how Theorem 2.7 and the Lemma 2.7b that follows from it can be exploited to determine coincidence of an eventuality from the eventuality sequence x (e.g. $x_p$) and another eventuality from y (e.g. $y_q$) within a cycle of the double recurrence (x, y) in time. It is an overview of how Algorithm 2 in section 3 works. It determines the coincidence of $x_p$ and $y_q$ by searching for an eventuality pair $w_r$ and $z_s$ from the gcd-partition (w, z) of the double recurrence (x, y), which are naturally non-disjoint with $x_p$ and $y_q$ respectively, such that if $w_r$ and $z_s$ co-occur over some subinterval k of any cycle, then it follows that the intervals of occurrence of $x_p$ and $y_q$ have a common subinterval which is non-disjoint with k.

The proof is concluded when it is realized that: there exists r within bounds $0 \leq r \leq len(w) - 1$ and s within bounds $0 \leq s \leq len(z) - 1$ such that:

$$v_p \equiv w_r + z_s. \quad \square$$

Proof of Theorem 2.8

*Let the pair of eventuality sequence (w, z) be a gcd-partition for the double recurrence eventuality sequence pair (x, y) and there exists an interval $\omega$ such that $\omega \in \Omega(x, y)$.

Let there exist r and s such that $0 \leq r \leq len(w) - 1$ and $0 \leq s \leq len(z) - 1$ such that

NN-Disjoint($x_p$, $w_r$) and NN-Disjoint($y_q$, $z_s$)

There exist k, $k_1$, j, m such that k, $k_1 \sqsubseteq \omega$ Occurs($x_p$, j) (where k and j are within the same interval of occurrence of x) that Occurs($y_q$, m) (where k and m are within the same interval of occurrence of y) and j, m $\sqsubseteq \omega$.

Because the statement marked * holds, then by Theorem 2.7, it will the case sometime within the interval $\omega$, that $k = k_1$.

If part

Let it be the case then that:

$\neg$Disjoint(common(k, j), common($k_1$, m))

In that case according to Axiom 1.0(a) there exists an interval n such that

n = common(common(k, j), common(k, m))

From this last conclusion and by Axiom 1.0 (c) there exists interval i such that

i = common(j, m).

Thus by Axiom 1.0 (a) it is the case that ¬Disjoint(j, m). Taking this with Occurs($x_p$, j) and Occurs($y_q$, m) and j, m $\sqsubseteq$ ω, then we can conclude from Definition 1.1 that Coincidence ($x_p$, $y_q$, ω).

<u>Only If part</u>

Let Coincidence ($x_p$, $y_q$, ω) for p such that $0 \leq p \leq len(x) -1$ and q such that $0 \leq q \leq len(y) -1$.
Let n be a common interval of subintervals j and m of ω such that Occurs($x_p$, j) and Occurs($y_q$, m).

By Lemma 2.7(b), there exists an interval such that k such that Occurs($w_r$, k) and Occurs($z_s$, k) which is non-disjoint with common(j, m). This means common(k, common(j, m)) exists.

By Theorem 1.0(d), it is the case that:

common(k, common(j, m)) = common(common(k, j), common(k, m))

Thus common(common(k, j), common(k, m)) exists from the fact that common(k, common(j, m)) exists. By Axiom 1.0(a), it follows that:

¬ Disjoint(common(k, j), common(k, m))

The proof will now be concluded by showing that:

k = k implies that ¬Disjoint(common(k, j), common(k, m))

Note that the left hand side of the implication statement is true and its right hand side has been inferred. Thus, the proof is concluded. □

<u>Proof of Theorem 2.9</u>

Let the pair of eventuality sequence (w, z) be a gcd-partition for an eventuality sequence pair (x, y) and
Let there exist r, s such that $0 \leq r \leq len(w) -1 \wedge 0 \leq s \leq len(z)-1$ such that

N-Subinterval($w_r$, $x_p$) or N-Subinterval($z_s$, $y_q$)

Reason by Cases:
Case1: Let N-Subinterval($x_p$, $w_r$).
By Axiom 2.3, there must exist some eventuality $z_s$, for NN-Disjoint($y_q$, $z_s$).
Let us consider any interval ω such that $\omega \in \Omega(x, y)$.
From the first assumption in the proof and Theorem 2.7, there exists an interval k such that j $\sqsubseteq$ ω and Occurs($w_r$, k) and Occurs($z_s$, k).

Because N-Subinterval($x_p$, $w_r$) and Definition 2.2b there exists an interval j, such that j⊑ω and Occurs($x_p$, j) and Subinterval(k, j).
Similarly, because NN-Disjoint($y_q$, $z_s$) and Definition 2.2a, there exists an interval m such that: m ⊑ ω and Occurs($y_q$, m) and Non-Disjoint(k, m).
(Note that j is within the same interval of occurrence of x that k is part of. Similarly, note that m is within the same interval of occurrence of y that k is part of. Thus, j and m are within the interval ω).

Thus: Subinterval(k, j) and Non-Disjoint(k, m) hold. Therefore, that Non-Disjoint(j, m).
Thus: Occurs($x_p$, j) and Occurs($y_q$, m) and Non-Disjoint(j, m) and j, m, k ⊑ ω. These facts constitute, according to Definition 1.1, the necessary and sufficient condition for defining Coincidence($x_p$, $y_q$, ω). Therefore coincidence of $x_p$ and $y_q$ within ω is proved.

Case 2: N-Subinterval($z_s$, $y_q$). For this case a proof can be constructed that is isomorphic to that of Case 1, such that $x_p$ and $y_q$ are exchanged and $w_r$ and $z_s$ are exchanged. □

Proof of Theorem 2.10
Let the pair of eventuality sequence (w, z) be a gcd-partition for an eventuality sequence pair (x, y) and

Let there exist r and s such that $0 \leq r \leq len(w) -2 \wedge 0 \leq s \leq len(z)-2$ such that
N-Overlaps($w_r$, $x_p$) and N-Overlaps($z_s$, $y_q$).

Let us consider any interval ω such that $\omega \in \Omega(x, y)$.

From the first assumption in the proof and Theorem 2.7, there exists an interval k such that k ⊑ ω and Occurs($w_r$, k) and Occurs($z_s$, k).

Because N-Overlaps($w_r$, $x_p$) and N-Overlaps($z_s$, $y_q$), then by the Definition 2.2c of N-Overlap:

> there exists two time intervals j and m such that Occurs($x_p$, j) and Occurs($y_q$, m) and Overlaps(k, j) and Overlaps(k, m). Note that j is within the same interval of occurrence of x that k is part of. Similarly, note that m is within the same interval of occurrence of y that k is part of. Thus, j and m are within the interval ω.

Thus, the boundaries of the common intervals of the pairs k and m as well as k and j denoted common(k, j) and common(k, m) can be define thus:

start( common(k, j)) = start(j) and end(common(k, j)) = end(k)
start(common(k, m)) = start(m) and end(common(k, m)) = end(k)

Thus the boundaries of the common interval of common(k, j) and common(k, m) is defined as:
end(common(common(k, j), common(k, m))) = end(k)
start(common(common(k, j), common(k, m))) = maximum-of(start(j), start(m))

The fact that: common(common(k, j), common(k, m)) exists means ¬Disjoint(j, m).

Thus a common interval exists between j and m which are the intervals of occurrences of $x_p$ and $y_q$ that occur within the interval ω, which is a cycle of the double recurrence (x, y), and its boundaries have been defined.

The facts: Occurs($x_p$, j) and Occurs($y_q$, m) and ¬Disjoint(j, m) and j, m ⊑ ω constitute according to Definition 1.1 the necessary and sufficient for defining Coincidence($x_p$, $y_q$, ω). Therefore coincidence between $x_p$ and $y_q$ is proved.□

Proof of Theorem 2.11

Let the pair of eventuality sequence (w, z) be a gcd-partition for an eventuality sequence pair (x, y).

**Let there be r and s such that $0 \leq r \leq len(w) - 1 \wedge 0 \leq s \leq len(z) - 1$ and

N-Subinterval($x_p$, $w_r$) and N-Subinterval($y_q$, $z_s$)

Let ω exist such that $\omega \in \Omega(x, y)$.

If Part

Let j, k, $j_1$, $k_1$ and ω exist such that k, $j_1$, $k_1$ ⊑ ω, and Occurs($w_r$, j) and any interval k of occurrence of $x_p$ which is not disjoint with j (within an interval of occurrence of x) and for any interval $j_1$ such that Occurs($z_s$, $j_1$) and any interval $k_1$ of occurrence of $y_q$ which is not disjoint with $j_1$ (within an interval of occurrence of y) and

$$\text{start}(k) - \text{start}(j) \leq \text{start}(k_1) - \text{start}(j_1) < \text{end}(k) - \text{start}(j) \text{ or}$$
$$\text{start}(k_1) - \text{start}(j_1) \leq \text{start}(k) - \text{start}(j) < \text{end}(k_1) - \text{start}(j_1).$$

By Theorem 2.7, there exists a subinterval of ω which is both an interval of occurrence of $w_r$ and $z_s$. Thus, let $j = j_1$, then it follows from the inequalities above that:

$$\text{start}(k) \leq \text{start}(k_1) < \text{end}(k) \text{ or}$$
$$\text{start}(k_1) \leq \text{start}(k) < \text{end}(k_1)$$

It follows that:

$$\text{start}(k) \leq \text{start}(k_1) < \text{end}(k) \text{ or}$$
$$\text{start}(k_1) \leq \text{start}(k) < \text{end}(k_1).$$

Both $\text{start}(k) \leq \text{start}(k_1) < \text{end}(k)$ and $\text{start}(k_1) \leq \text{start}(k) < \text{end}(k_1)$ imply that k and $k_1$ are non-disjoint i.e. ¬Disjoint(k, $k_1$) by Definition 2.5. if we add to this fact, the facts that:

Occurs($x_p$, k) and Occurs($y_q$, $k_1$) and k, $k_1$ ⊑ ω.

The necessary and sufficient condition for inferring Coincidence($x_p$, $y_q$, ω), according to Definition 1.1 has been proved. Thus, coincidence between $x_p$ and $y_q$ exists within the interval ω.

Only If Part

Let a coincidence occur between $x_p$ and $y_q$ within the interval ω.
By Theorem 2.7, there is a subinterval of ω which is an interval of occurrence for both $w_r$ and $z_s$.

For the situation in which Occurs($x_p$, k) and Occurs($w_r$, j) such that k and j are part of the same occurrence of x, and Occurs($y_q$, $k_1$) and Occurs($z_s$, $j_1$) such that $k_1$ and $j_1$ are part of the same occurrence of y, and j, k, $j_1$, $k_1$ $\in \Omega$.

From the statement marked ** and the Definition 2.2(c) of N-Subinterval, it is the case that Subinterval(k, j) and Subinterval($k_1$, $j_1$).

Let $j = j_1$, then

1. The common interval of k and j will have the same boundaries as k, and the common interval of $k_1$ and $j_1$ will have the interval the same boundaries as $k_1$.

2. By Theorem 2.5, k and $k_1$ are disjoint if and only if:
   $\text{start}(k) \leq \text{start}(k_1) < \text{end}(k)$ or
   $\text{start}(k_1) \leq \text{end}(k) < \text{end}(k_1)$

3. Because $j = j_1$, the disjunction of inequalities above can be re-written as:
   $\text{start}(k) - \text{start}(j) \leq \text{start}(k_1) - \text{start}(j_1) < \text{end}(k) - \text{start}(j)$ or
   $\text{start}(k_1) - \text{start}(j_1) \leq \text{start}(k) - \text{start}(j) < \text{end}(k_1) - \text{start}(j_1)$.

The proof is concluded. □

Proof of Theorem 2.12
Let the pair of eventuality sequence (w, z) is a gcd-partition for an eventuality sequence pair (x, y).

Let an interval ω exist such that $\omega \in \Omega(x, y)$.

*Let there exists particular r and s such that $0 \leq r \leq \text{len}(w)\text{-}1 \wedge 0 \leq s \leq \text{len}(z)\text{-}2$ such that
N-Subinterval($x_p$, $w_r$) and N-Overlaps($z_s$, $y_q$) and ¬N-Subinterval($z_{s+1}$, $y_q$)

**The facts that N-Overlaps($z_s$, $y_q$) and ¬N-Subinterval($z_{s+1}$, $y_q$) imply N-Overlaps($y_q$, $z_{s+1}$).

If Part
Let j, k, $j_1$, $k_1$ and ω exist such that k, $j_1$, $k_1$ $\sqsubseteq \omega$, and Occurs($w_r$, j) and any interval k of occurrence of $x_p$ which is not disjoint with j (within an interval of occurrence of x) and for any interval $j_1$ such that Occurs($z_s$, $j_1$) and any interval $k_1$ of occurrence of $y_q$ which is not disjoint with $j_1$ (within an interval of occurrence of y).

Furthermore, let the interval of occurrence of $z_{s+1}$ that meets $j_1$ be denoted by $succ(j_1)$ where succ is some kind of succession function for intervals of occurrence for eventualities from gcd partitions. Thus, $end(j_1) = start(succ(j_1))$.

Note that from the statements marked * and ** and the Definitions 2.2(b and c) as well as the facts that Occurs($w_r$, j) and Occurs($x_p$, k) that Subinterval(k, j). Similarly, the facts that Occurs($z_s$, $j_1$) and Occurs($w_r$, $k_1$) and Occurs($z_{s+1}$, succ($j_1$)), it is the case that Overlaps($j_1$, $k_1$) and Overlaps($k_1$, succ($j_1$)).

Thus let it be the case that:

$$start(k) - start(j) \leq start(k_1) - start(j_1) < end(k) - start(j) \text{ or}$$
$$start(k_1) - start(j_1) \leq start(k) - start(j) < end(j_1) - start(j_1) \text{ OR}$$
$$0 \leq start(k) - start(j) < end(k_1) - end(j_1).$$

By Theorem 2.7, for every pair $w_r$ and $z_s$ there exists a subinterval of $\omega$ which is an interval of occurrence for both $w_r$ and $z_s$ and another subinterval of $\omega$ which is an interval of occurrence for both $w_r$ and $z_{s+1}$.

Let us consider what happens within those two intervals. The first interval is the interval over which the occurrence of $w_r$ and $z_s$ are both the same interval i.e. $j = j_1$.

In that case, the disjunction of inequalities can be re-written:

$$start(k) \leq start(k_1) < end(k) \text{ or}$$
$$start(k_1) \leq start(k) < end(j_1) \text{ OR}$$
$$0 \leq start(k) - start(j) < end(k_1) - end(j_1)$$

The first two disjuncts taken together means that the interval k and the interval with boundaries start($k_1$) and end ($j_1$) are not disjoint. The latter interval is the common subinterval of $j_1$ and $k_1$, which have the relationship Overlaps($j_1$, $k_1$). Therefore, we can conclude that:

$$\neg Disjoint(k, common(j_1, k_1)).$$

Bearing in mind that $end(j_1) = start(succ(j_1))$, the last inequality in the disjunction can be re-written as:

$$start(succ(j_1)) - start(succ(j_1)) \leq start(k) - start(j) < end(k_1) - start(succ(j_1))$$

In the second case, if $j = succ(j_1)$ then

$$start(succ(j_1)) \leq start(k) < end(k_1).$$

This means that interval k is non disjoint with an interval with boundaries at start(succ($j_1$)) and end($k_1$). The interval with these boundaries start(succ($j_1$)) and end($k_1$) is the common interval of $k_1$ and succ($j_1$), which have the relationship Overlaps($k_1$, succ( $j_1$)). Therefore, we can conclude that:

$$\neg Disjoint(k, common(k_1, succ(j_1))).$$

Combining the outcomes of the two cases, it is the case that

$$\neg\text{Disjoint}(k, \text{common}(j_1, k_1)) \text{ or} \neg\text{Disjoint}(k, \text{common}(k_1, \text{succ}(j_1))).$$

This implies that k and $k_1$ are non-disjoint, i.e. $\neg$Disjoint(k, $k_1$)

If we combine this new fact with the facts that k, $k_1 \sqsubseteq \omega$, and that Occurs($x_p$, k) and Occurs($y_q$, $k_1$), the necessary and sufficient conditions for defining Coincidence($x_p$, $y_q$, $\omega$) according to Definition 1.1 have been fulfilled. The If-part of the proof has been concluded.

Only If Part

Assume that there is a coincidence of $x_p$ and $y_q$ within $\omega$.

Let j, k, $j_1$, $k_1$ and $\omega$ exist such that k, $j_1$, $k_1 \sqsubseteq \omega$, and Occurs($w_r$, j) and any interval k of occurrence of $x_p$ which is not disjoint with j (within an interval of occurrence of x) and for any interval $j_1$ such that Occurs($z_s$, $j_1$) and any interval $k_1$ of occurrence of $y_q$ which is not disjoint with $j_1$ (within an interval of occurrence of y).

Let it be the case that Occurs($z_{s+1}$, succ($j_1$)) as defined in the If-part of the proof, so that end($j_1$) = start(succ($j_1$)).

By Theorem 2.8, the common intervals of the intervals of occurrence of both $x_p$ and $y_q$ must share a common subinterval with the intervals of occurrence of $w_r$ and $z_s$ (or $z_{s+1}$).

That can only happen If k and $k_1$ will share a common subinterval with j when either j = $j_1$ or j = suc($j_1$), that common subinterval must be a subinterval of either $j_1$ or succ($j_1$).

Case 1: Let j = $j_1$.

By Theorem 2.7, there exists a subinterval of $\omega$ which is both an interval of occurrence of $w_r$ and $z_s$. Thus, let j = $j_1$. Then it follows from the fact that N-Overlaps($z_s$, $y_q$) and N-Subinterval($x_p$,$w_r$) and Definition 2.2(b and c) of N-Overlaps and N-Subinterval, that Overlaps($j_1$, $k_1$) and Subinterval(k, j).

Thus, if j = $j_1$ then Overlaps(j, $k_1$) and Subinterval(k, j). The boundaries of the common interval of j and $k_1$ are defined thus:

$$\text{start}(\text{common}(j, k_1)) = \text{start}(k_1)$$
$$\text{end}(\text{common}(j, k_1)) = \text{end}(j)$$

There is coincidence between k and $k_1$ that is within $j_1$ if:

$$\text{start}(k_1) \leq \text{start}(k) < \text{end}(j_1) \text{ or}$$
$$\text{start}(k) \leq \text{start}(k_1) < \text{end}(k).$$

Since j = $j_1$, this disjunction of inequalities can be re-written as:

$$\text{start}(k_1) - \text{start}(j_1) \leq \text{start}(k) - \text{start}(j) < \text{end}(j_1) - \text{start}(j_1) \text{ or}$$
$$\text{start}(k) - \text{start}(j) \leq \text{start}(k_1) - \text{start}(j_1) < \text{end}(k) - \text{start}(j)$$

The proof is concluded for this case.

Case 2: Let $j = succ(j_1)$

Similarly, by the same Theorem 2.7, there exists a subinterval of ω which is both an interval of occurrence for both $w_r$ and $z_{s+1}$. Thus, let $j = succ(j_1)$. Then it follows from the facts that N-Overlaps($y_q$, $z_{s+1}$) and N-subinterval($x_p$, $w_r$), that Overlaps($k_1$, succ($j_1$)) and Subinterval(k, j).

Thus, if $j = succ(j_1)$, then Overlaps($k_1$, succ($j_1$)) and Subinterval(k, succ($j_1$)). The boundaries of the common interval of $k_1$ and succ($j_1$) are defined thus:

$$start(common(k_1, succ(j_1)) = start(succ(j_1))$$
$$end(common(k_1, succ(j_1)) = end(k_1).$$

There is coincidence between k and $k_1$ within succ($j_1$) if :

$$start(succ(j_1)) \leq start(k) < end(k_1).$$

Note that $start(succ(j_1)) = end(j_1)$. Thus, the inequalities above can be re-written as:

$$end(j_1) \leq start(k) < end(k_1)$$

Similarly, since in this case, since in this case, $j = succ(j_1)$, then,

$$end(j_1) – start(succ(j_1)) \leq start(k) – start(j) < end(k_1) – start(succ(j_1)))$$

because $end(j_1) = start(succ(j_1))$ then this inequality can be re-written as:

$$0 \leq start(k) – start(j) < end(k_1) – end(j_1)$$

Now let us determine the boundaries of the common intervals for the pairs of intervals arrived at in the last two steps. Combining the outcomes of Cases 1 and 2 leads to the disjunction of inequalities:

$$start(k) – start(j) \leq start(k_1) – start(j_1) < end(k) – start(j) \text{ or}$$
$$start(k_1) – start(j_1) \leq start(k) – start(j) < end(j_1) – start(j_1) \text{ OR}$$
$$0 \leq start(k) – start(j) < end(k_1) – end(j_1) .$$

The proof is concluded for the second case. □

Proof of Theorem 3.1

In terms of running time, lines 4 and 5 are the only lines that require linear time running time. Line 4 has a running time of O(len(x)) while line 5 has a running time of O(len(y)).

The maximum number of eventualities (i.e. value of len(x)) that can possibly be in x is duration(x).

Similarly, the maximum number of eventualities (i.e. value of len(x)) that can possibly be in y is duration(y).

Every other statement in the algorithm will run in constant time.

The running time of the algorithm is duration(x) + duration(y) + O(1).

Thus the running time is O(max(duration(x), duration(y))).□